\documentclass[runningheads,dvipsnames]{llncs}
\usepackage[T1]{fontenc}
\usepackage{graphicx}
\usepackage{import}
\usepackage{amssymb}
\usepackage{amsmath}
\usepackage{booktabs}
\usepackage{caption}
\usepackage{subcaption}
\usepackage{url}
\usepackage{tabularx}
\newcolumntype{Y}{>{\centering\arraybackslash}X}
\newcolumntype{Z}{>{\raggedleft\arraybackslash}X}
\usepackage{multirow}
\usepackage[table]{xcolor}
\definecolor{AccRow}{gray}{1.0}       
\definecolor{AdvAccRow}{gray}{0.95}    
\definecolor{TimeRow}{gray}{0.85}      
\usepackage[linesnumbered,ruled,vlined]{algorithm2e}

\usepackage{xcolor}
\usepackage{ulem}
\colorlet{macolor}{violet!80}
\colorlet{bgcolor}{olive!80}
\definecolor{hhcolor}{rgb}{0.2,0.6,0.6}
\colorlet{delcolor}{black!50}
\colorlet{todocolor}{RedOrange!100}
\colorlet{changedcolor}{WildStrawberry!100}
\newcommand{\nbc}[3]{
		{\colorbox{#3}{\bfseries\sffamily\scriptsize\textcolor{white}{#1}}}
		{\textcolor{#3}{\sf\small$\blacktriangleright$\textit{#2}$\blacktriangleleft$}}
}

\newcommand{\ma}[1]{\nbc{MA}{#1}{macolor}}
\newcommand{\hh}[1]{\nbc{HH}{#1}{hhcolor}}
\newcommand{\bg}[1]{\nbc{BG}{#1}{bgcolor}}
\newcommand{\todo}[1]{\nbc{TODO}{#1}{todocolor}}
\newcommand{\changed}[1]{\textcolor{changedcolor}{$\blacktriangleright$ #1 $\blacktriangleleft$}}
\newcommand{\deleted}[1]{\nbc{DEL}{\sout{#1}}{delcolor}}

\newcommand{\hide}[1]{}

 \renewcommand{\changed}[1]{#1}
 \renewcommand{\deleted}[1]{}
 \renewcommand{\todo}[1]{}
 \renewcommand{\ma}[1]{}
 \renewcommand{\hh}[1]{}
 \renewcommand{\bg}[1]{}

\usepackage{cleveref}

\usepackage{ragged2e}
\begin{document}
\title{On the Efficiency of Training \linebreak{}Robust Decision Trees}
%
%
\author{Benedict Gerlach\inst{1} \and
Marie Anastacio\inst{1,2}\orcidID{0000-0002-4039-2470} \and
Holger H. Hoos\inst{1,2,3}\orcidID{0000-0003-0629-0099}}
\authorrunning{B. Gerlach et al.}
%
\institute{
     AIM, RWTH Aachen University, Aachen, Germany
    \and LIACS, Leiden University, Leiden, The Netherlands
    \and University of British Columbia (UBC), Vancouver, Canada
}
\maketitle              
%
\begin{abstract}

As machine learning gets adopted into the industry quickly, trustworthiness is increasingly in focus. Yet, efficiency and sustainability of robust training pipelines still have to be established. In this work, we consider a simple pipeline for training adversarially robust decision trees and investigate the efficiency of each step. 
Our pipeline consists of three stages.
Firstly, we choose the perturbation size automatically for each dataset. For that, we introduce a simple algorithm, instead of relying on intuition or prior work. 
Moreover, we show that the perturbation size can be estimated from smaller models than the one intended for full training, and thus significant gains in efficiency can be achieved.
Secondly, we train state-of-the-art adversarial training methods and evaluate them regarding both their training time and adversarial accuracy.
Thirdly, we certify the robustness of each of the models thus obtained and investigate the time required for this. We find that verification time, which is critical to the efficiency of the full pipeline, is not correlated with training time.

\keywords{verification \and decision trees \and efficiency \and trustworthiness \and adversarial robustness \and explainability \and machine learning}
\end{abstract}
\section{Introduction}
As more individuals, companies, and governments incorporate AI into their work patterns and processes, we need it to be \textbf{trustworthy} and \textbf{compute efficient}. 
Trustworthiness has many definitions. It is commonly associated with 
\hh{slightly modified:}
a machine learning model or system being accurate, but also explainable, since it allows one to pinpoint potential errors and their origin.  Compute efficiency is also required to render machine learning financially and environmentally acceptable.
Despite the rise of deep learning, those needs allowed decision trees (DTs), ensembled DT models such as random forests (RF)~\cite{breiman_random_2001}, and gradient-boosted trees (GBT)~\cite{prokhorenkova_catboost_2018,chen_xgboost_2016} to remain prominent choices, especially for sensitive applications, e.g., in healthcare~\cite{vlachas_random_2022,loef_using_2022} and finance~\cite{hamza_defd_2023}, which involve working with tabular data and in low data regimes. The simple linear structure of these models in combination with efficient training and inference procedures allow them to meet the previously stated requirements well.

However, similarly to other machine learning models~\cite{langenberg_robustness_2019,sitawarin_robustness_2019}, including deep neural networks~\cite{szegedy_intriguing_2014,goodfellow_explaining_2015,carlini_adversarial_2017,madry_towards_2018},
these simpler models are susceptible to adversarial attacks~\cite{kantchelian_evasion_2016,xu_automatically_2016}.
To counteract this problem, adversarial defences have been developed~\cite{chen_robust_2019,chen_robustness_2019,vos_efficient_2021}, largely through adversarial training. These methods and the verification of their correctness should be efficient as well.
However, training models robustly requires more computing resources than traditional training, which puts trustworthiness and efficiency at odds with each other.

We present an empirical efficiency analysis of the components of a simple training pipeline for robust DTs. 
Robustness is typically defined depending on the allowed size of perturbations on the input~\cite{szegedy_intriguing_2014,kantchelian_evasion_2016}, which has to be decided before any training is done. This is the first step of our pipeline, in which we present a method to select an appropriate perturbation size for each given dataset. We investigate the impact of model size on the perturbation size, thus determined and analyse whether the perturbation size can be selected using smaller models than the intended ones for efficiency purposes. Our findings indicate a possible efficiency improvement of several orders of magnitude, although additional research is needed to unlock this potential. 

The second step is to train the model. We optimise and evaluate various defensive training strategies for DT ensembles, keeping track of their accuracy, adversarial accuracy and training time. We find that Groot RFs~\cite{vos_efficient_2021} and Robust RFs~\cite{chen_robust_2019} clearly outperform others with respect to the (adversarial) accuracy achieved, given their training time efficiency. 

The last step is to verify the robustness of the trained model. We evaluate the impact of earlier decisions on the verification time of the resulting models using the main solver of the open-source optimisation suite SCIP~\cite{bolusani_scip_2024} to solve the mixed-integer linear program (MILP) formulation by Kantchelian {\it et al.}~\cite{kantchelian_evasion_2016}.
We find that well-performing methods with good training time efficiency typically have large verification times, and the converse also holds. Given that our pipeline uses hyperparameter optimisation (HPO) and validates the robustness of models in that phase, verification time has a large influence on pipeline efficiency.

In the following, we first describe the background and related work on adversarial attacks, introducing our notation and basic concepts. 
Analogous to the three steps of our pipeline, the experimental section is partitioned into three sections, each discussing one stage and the results of our experiments. We then comment on how those stages interact and draw conclusions.

\section{Background and Related Work}
\label{sec:background}

Models trained with standard methods often perform poorly under adversarial attacks, even under small perturbations~\cite{kantchelian_evasion_2016,xu_automatically_2016,chen_robust_2019,vos_efficient_2021}, which motivated the development of many methods to train models such that they are robust to those attacks. However, models trained to be adversarially robust typically achieve lower  accuracy~\cite{zhang_theoretically_2019} and require longer training time~\cite{chen_robust_2019,vos_efficient_2021} when compared to standard models on the same dataset. 

\subsection{Adversarial attacks}

Given a classifier $f$ trained on a dataset $D$ and a input $x\in D$, an adversarial attack is a perturbed input $\tilde{x}=x+\delta$ that 
causes a misclassification $f(\tilde{x})\neq f(x)$~\cite{szegedy_intriguing_2014,kantchelian_evasion_2016}. The size of the perturbation is $\varepsilon=\lvert \delta \rvert_p $, given some norm $p$.
Following Yang {et al.}~\cite{yang_certified_2021}, the model $f$ is called $\varepsilon$-robust on $x$, if 
$\forall \delta: \lvert \delta \rvert_p \leq \varepsilon, f(x)=f(x+\delta)$. 
\hh{modified:}
We define the adversarial accuracy of $f$ on dataset $D$ under $\varepsilon$, noted $\textrm{AdvAcc}_f^D(\varepsilon)$, as the fraction of samples on which $f$ is $\varepsilon$-robust.

\subsection{Choosing Perturbation Size and Threat Model}
\label{sec:back/pert}
Both training and verification methods require selecting perturbation sizes; these do not necessarily have to be identical, although they are typically chosen that way~\cite{chen_robust_2019,andriushchenko_provably_2019,vos_efficient_2021}. Also, the $p$-norm, more broadly a distance measure, has to be chosen to fit the threat scenario~\cite{andriushchenko_provably_2019,calzavara_treant_2020,xu_towards_2022}. Because it is the most studied, we focus on the $\ell_\infty$ norm~\cite{chen_robust_2019,vos_efficient_2021,andriushchenko_provably_2019}. Extensions to arbitrary distances are possible and are currently being studied~\cite{xu_towards_2022,kireev_adversarial_2023,chen_cost-aware_2021,calzavara_treant_2020,kulynych_evading_2018}, especially in the context of tabular data. However, the perturbation sizes have, to the best of our knowledge, been chosen arbitrarily for each dataset.

To choose these sizes algorithmically, we introduce the adversarial success rate $\eta$, the relative gap \changed{between the accuracy of $f$ on $D$ ($\text{Acc}^D_f$) and the respective adversarial accuracy ($\text{AdvAcc}_f^D(\varepsilon)$) given attack size $\varepsilon$}\deleted{in the accuracy of $f$ on $D$, caused by a given attack size}:
\begin{align}
    \eta (\varepsilon) = 1 - \frac{\textrm{AdvAcc}_f^D(\varepsilon)}{\textrm{Acc}^D_f},
\end{align}
\label{eq:advloss}
We aim to select an $\hat{\varepsilon}$ that induces a certain success rate goal $\eta^*$.

\subsection{Adversarial Training of a Model}
\label{sec:back/train}

To render models more robust, many adversarial training methods have been proposed~\cite{andriushchenko_provably_2019,andriushchenko_understanding_2020,calzavara_treant_2020,chen_robust_2019,vos_efficient_2021}.
However, models trained to be adversarially robust typically perform worse in classification {than models trained on the same datasets using standard methods}.
\hh{slightly modified:}
This is known as the ``accuracy Gap'' \cite{tsipras_robustness_2018,chen_robust_2019,vos_efficient_2021}; it has been shown that there is a trade-off between minimizing this gap and maximizing robustness~\cite{zhang_theoretically_2019}.

Adversarial training methods can be categorised into three groups, depending on their mechanism and the types of models they can be applied to:
\begin{enumerate}
    \item Robust splitting criteria: These incorporate some notion of easy-to-evade samples into the individual splitting loss. They thus can be applied both to single trees and Random Forests (RFs).
    \item Robust boosting: These modify the boosting loss to incorporate robustness as a goal that needs to be maximized similarly to accuracy. New weak classifiers are thus constructed to improve the robustness of the ensemble.
    \item Noisy training data: These add noise to the clean data with the goal of increasing robustness. They can be applied to all types of models.
\end{enumerate}
Robust splitting and boosting can incorporate exact solutions, which require a lot of computation and can become infeasible to train, thus leading to the development of heuristic approaches. 
\hh{modified -- check:}\bg{added time complexity for adv boosting methods}
For each of the following methods, we state their time complexity; this only concerns the additional work required for considering the adversarial scenario. 

Groot~\cite{vos_efficient_2021} is a robust splitting criterion. It takes optimal local splitting decisions whilst only requiring constant time to consider all split options in a feature. 
Since it uses a robust splitting criterion to create a single tree, it can be applied to random forests as well. 
A similar algorithm running in log-time was introduced earlier by Chen {\it et al.}~\cite{chen_robust_2019}, a heuristic version of which is also running in constant time. In the remainder of this paper, we refer to the RFs trained with this constant time heuristic as Robust RFs.
Treant~\cite{calzavara_treant_2020} uses another adversarial splitting formulation, but with a loss-based splitting approach, that requires more time, due to iterative loss minimization.

For Gradient-boosted decision trees only approximative and heuristic log-time solutions exist; exact methods would be too costly. Our experiments include provably robustly boosted decision stumps (Robust Boost)~\cite{andriushchenko_provably_2019} and robust gradient boosted decision trees (Robust Trees)~\cite{chen_robust_2019}, which use an adapted version of the constant time heuristic for Robust RFs. Robust Boost requires an additional factor of $\mathcal{O}(\log T \log n)$ and Robust Trees no additional complexity compared to standard GBDT constructed via XGBoost, $T$ is the number of trees and $n$ the number of samples.

Noisy data methods, as we use them, are undirected. Directed adversarial data generation, sometimes used for neural networks~\cite{goodfellow_explaining_2015,wong_fast_2020}, is redundant, since other methods implicitly solve splitting for such data. Random forests created with this method are denoted as Noisy RF and used as a baseline in the following.

\subsection{Verification of Adversarial Robustness}
\label{sec:back/verif}


Adversarial robustness has to be verified for large sets of inputs in order to ensure the robustness of a given model.
The robustness of state-of-the-art decision trees can be verified exactly via MILP~\cite{kantchelian_evasion_2016} and satisfiability modulo theories (SMT)  solvers~\cite{nie_varf_2020} or by bounding the input space directly~\cite{tornblom_abstraction-refinement_2019,matsunaga_efficient_2024}, or heuristically with~\cite{chen_robustness_2019} or without bounds~\cite{zhang_efficient_2020}. Solving the adversarial robustness of a sample to optimality requires the identification of a worst-case attack, a perturbation of minimal size that leads to misclassification. Given a perturbation of size $\varepsilon$, the optimality problem becomes a feasibility one, which is potentially much easier. Despite large computational cost, both are generally possible for DTs, unlike exact verification of state-of-the-art neural networks~\cite{huang_survey_2020}. {Nonetheless,} efficient choices are important throughout the training pipeline to render the entire process tractable.
\hh{to make (better: render) what tractable?}\bg{the entire training pipeline as a whole}


Depending on the defence method used, trees are constructed differently, potentially aiding or counteracting fast exact verification. 
\hh{modified:}
Obviously, model size is a relevant factor, and we focus on it in \Cref{sec:exp/pert}, but split selection during training shapes the bounded polyhedron, which is the solution space for SMT problems and MILP~\cite{kantchelian_evasion_2016,nie_varf_2020}. The choice of the best adversarial training method should incorporate knowledge of the time required for verifying the corresponding models.

\section{Setup of experiments}
\label{sec:exp}

As indicated in \Cref{sec:background}, we separate the process of producing efficient adversarially robust decision trees into three stages
. In practice, these will influence one another, but for simplification, we consider them independently. 
Our implementation and supplementary plots and data are freely available.\footnote{www.github.com/ADA-research/DT-efficiency}

\subsubsection{Datasets.}
\label{sec:exp/data}
In the following sets of experiments, we use the datasets \textit{Wine, Ionosphere, German Credit Data \textnormal{(GCD)}, Adult, MS Malware Prediction, IEEE CIS Fraud Detection}. The first four are taken from the UCI Machine Learning Repository \footnote{archive.ics.uci.edu/} and the latter two from Kaggle. They span different sizes, complexities and types, as seen in \Cref{table:datasets}.
\begin{table}[tb]
\caption[Table categorizing used Datasets by Size, Type, Complexity, Class Count]{The  datasets used in our study are categorised by the number of samples they contain; 
\hh{modified -- check:}
the type of samples, where `structured' means only decimal values; the total feature complexity in Bits; and the number of target classes to predict.}
\label{table:datasets}
\centering
\begin{tabular}{rcrclcrcc}
\toprule
Dataset                 & & \#Samples& & Type       & & Complexity & &\#Classes \\
\midrule
Wine                    & & 178               & & Structured & & <\hspace{1pt} 100 B              & & 3  \\
Ionosphere              & & 351               & & Structured & & <\hspace{1pt} 100 B              & & 2\\
GCD                     & & 1000              & & Tabular    & & <1000 B            & & 2\\
Adult                   & & $\geq$10000            & & Tabular    & & <\hspace{1pt} 100 B              & &  2\\
MS Malware Prediction   & & $\geq$10000            & & Tabular    & & <1000 B              & & 2\\
IEEE CIS Fraud Detection& & $\geq$10000            & & Tabular    & & >1000 B        & & 2\\
\bottomrule
\end{tabular}
\end{table}
For bigger datasets, we limit the size to 10\,000 samples, and if possible sample from the dataset such that the target class ratio approaches the log ratio. By this, we mean that an original ratio of $e^2:e$ would approach $2:1$.

Categorical features were converted to numerical ones via ordinal encoding. 
This domain-independent transformation loses equal distance representation, mainly, for non-binary or non-ternary features, 
which {occur seldomly.}
For high-dimensional features, the possible equal distance among all classes is converted to a line with varying distances among classes.
Other representations could be investigated in the future, such as complex distances applicable to tabular data (see, {\it e.g.},~\cite{xu_towards_2022,kireev_adversarial_2023,chen_cost-aware_2021,calzavara_treant_2020,kulynych_evading_2018}).

\subsubsection{Procedure.}

All experiments were conducted independently on all datasets, with each set split into a training set, validation set and test set according to a ratio of 64:20:20. The test-set split was fixed across repetitions.
The verification computations were done with the method by Kantchelian {\it et al.}~\cite{kantchelian_evasion_2016}, 
\hh{modified -- check:}
using the SCIP solver, and the running times thus expended were recorded.
A maximum of 48 hours was spent on training, hyperparameter optimisation and verification for each method on each dataset. Some methods, mostly Treant, but also Noisy RF, could not finish within this time on the larger datasets Adult, IEEE CIS Fraud Detection, and MS Malware Prediction, and were thus excluded from the results on those datasets. On these datasets, when methods ran out of time in a small number of repetitions, we included them.
Memory was never exhausted during any of the runs. The experiments were repeated 7 times.
When we report results on example datasets, all other results can be found in supplementary material. 

\subsubsection{Execution environment.}
\label{sec:exp/env}
Our experiments were conducted on a high-performance compute cluster running Rocky Linux OS version 9.3.  Each node was equipped with two AMD EPYC 7543 CPUs with 32 cores and 256 MB of cache each as well as 1 TB of RAM. For each run, we used a single core.

\section{Impact of Model Size on Perturbation Size Search}
\label{sec:exp/pert}

The choice of attack size considered in our experiments (cf. \Cref{sec:back/pert}) should be such that it incurs a large loss in accuracy while being hard to detect and thus as small as possible. Such a size will depend on the method used and the dataset, see \Cref{table:epsilons}. However, for comparability in the latter sections, we need the attack sizes to be the same for all models. Thus, for this step, we optimise this size on a simple RF, which is our baseline and, contrary to other studied methods, does not depend on the attack size we are optimising.
Note that the sizes used in training and verification do not need to be the same; we only search for perturbation sizes for verification, while the perturbation size 
for training is optimised alongside the model hyperparameters.

\subsection{Search Algorithm}
\label{sec:exp/pert/algo}
The general algorithm involves 6 steps and is shown in \Cref{alg:basic_search}.
\begin{algorithm}
\caption{Iterative Perturbation Size Estimation}
\label{alg:basic_search}
\KwIn{Target success rate $\eta^*$, initial attack size estimate $\hat{\varepsilon}$}
\KwOut{Estimated $\hat{\varepsilon}$ such that $\eta(\hat{\varepsilon}) = \eta^*$}

Train surrogate model (e.g., random forest)\;
\Repeat{convergence of $\hat{\varepsilon}$ or $\eta(\hat{\varepsilon})$}{
    Evaluate attack feasibility on the test set using current $\hat{\varepsilon}$\;
    Partition test set into feasible and infeasible samples\;
    Compute adversarial success rate $\eta(\hat{\varepsilon})$\;
    Estimate $\hat{\varepsilon}$ such that $\eta(\hat{\varepsilon}) = \eta^*$\;
}
\Return{$\hat{\varepsilon}$}
\end{algorithm}

{We start by training the \changed{surrogate RF model}. Then, we evaluate if an attack can be made on a test set given an attack size $\hat{\varepsilon}$. We divide the set into two sets, the samples on which an attack was found and the rest, which allows us to compute the adversarial success rate for this attack size. Lastly, we estimate a new value for $\hat{\varepsilon}$ which would likely lead to $\eta(\hat{\varepsilon})=\eta^*$, our target success rate. We repeat this process iteratively until convergence or until we reach the target.}

\hh{slightly modified -- check carefully (was awkward):}
For efficiency, we maintain a distance array that stores, for each input from the test set, the lowest perturbation size discovered so far that results in misclassification. 
The evaluation in each round thus only needs to occur on the portion of inputs whose stored perturbation size is directly larger than the current estimate. Because this set gets refined 
over time, the additional time for each step is logarithmic, but can behave linearly. The next guess for a size is computed via a Newton-Raphson method using the success goal as an offset to the success rate function $\eta$. After a certain number of steps, the implementation falls back to a linear 
interpolation given the two best guesses of $\hat{\varepsilon}$ leading to success rates bracketing $\eta^*$. 
Efficiency is further increased by caching the perturbed samples with the smallest perturbation size 
found so far for each given input; this is used to initialise further checks on that input. 
Furthermore, SCIP was configured to solve for feasibility more optimally; we deactivated presolvers, employed feasibility pumping and set an emphasis on feasibility checking.


\subsection{Setup}
\label{sec:exp/pert/setup}

\hh{reworded -- check carefully:}
Using Algorithm 1, we determine a perturbation size $\hat{\varepsilon}$ roughly resulting in an 
adversarial success rate $\eta^*=0.1$.
The model used for computing the adversarial success rate $\eta$ is not trained adversarially, as this would create a dependence on the $\varepsilon$ we are searching. We use random forests as a baseline and thus select the perturbation distance depending on their vulnerability. We train on 80\% of a given dataset and keep the remaining 20\% for testing.


\hh{modified -- check carefully:}
Without requiring the use of dedicated methods for detecting attacks, our approach allows us to adapt attack size to a given dataset.
Moreover, we aim for a relatively low adversarial success rate, since large perturbations are likely to be flagged by detectors. Recent studies show detection rates above 80\% for vision datasets and up to 98\% for tabular datasets, with few false positives (see, {\it e.g.}, \cite{feldsar_detecting_2023,goodfellow_explaining_2015}). Furthermore, adversarial training methods should naturally increase robustness in low attack size regimes most.\footnote{Note that this is not always necessarily the case; e.g., Vos {\it et al.}\cite{vos_efficient_2021} created a method that puts equal weights on all possible inputs for which incorrect classification can be obtained
given a perturbation size.} 

Since we use only tabular or tabular-like data, we conservatively chose $\eta^*=10\%$ with corresponding perturbation size as a robustness target for the model. 
Moreover, we investigate the effects of RF size, in terms of depth and number of trees, on the resulting $\hat{\varepsilon}$ value and total time spent on verification. 
The size hyperparameters are drawn from a logarithmic 
grid, tree depth from $\{3,4,5,7,9\}$ and number of trees from $\{5,11,25,56,125\}$. 
The search algorithm is initialised based on preliminary solutions found with the smallest model, to avoid a {computationally expensive} 
search on larger models.
The success rate is set to converge within 2\% precision around the 10\% success goal. 
Because of limited numerical accuracy, we also {terminate the search}
if the two considered sizes are within a margin of about $10^{-6}$.
\hh{modified -- check:}
All runs were limited to a maximum of 12 hours wall-clock time.

\subsection{Results}
\label{sec:exp/pert/results}
\begin{figure}[t]
    \centering
    \begin{subfigure}{0.49\linewidth}
        \centering
        \includegraphics[width=\linewidth]{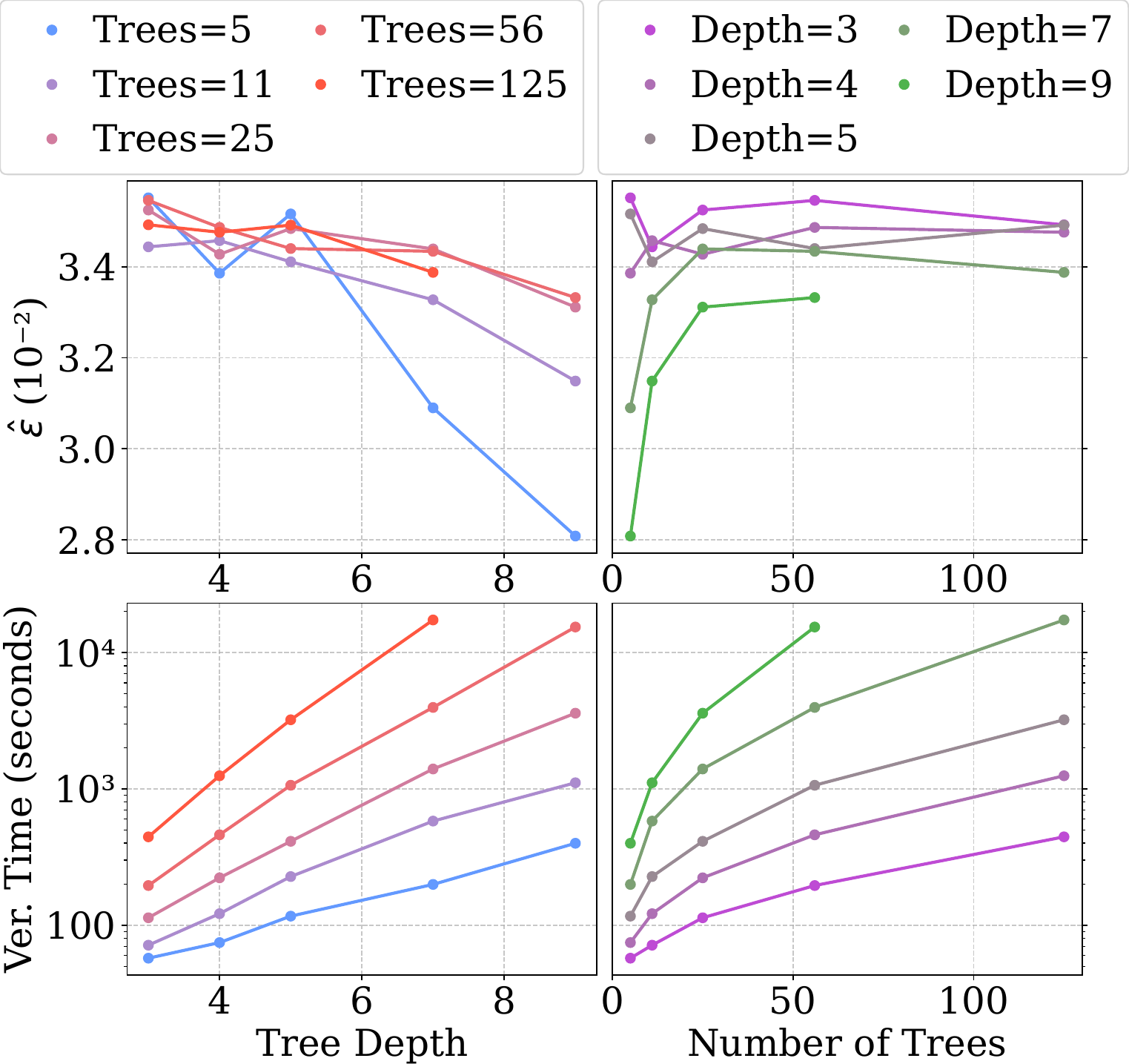}
    \subcaption[Adult Perturbation Size Search Result and Verification Time over Model Size]{on dataset Adult}
    \label{fig:size-medians/adult}
    \end{subfigure}
    \begin{subfigure}{0.49\linewidth}
    \centering
    \includegraphics[width=\linewidth]{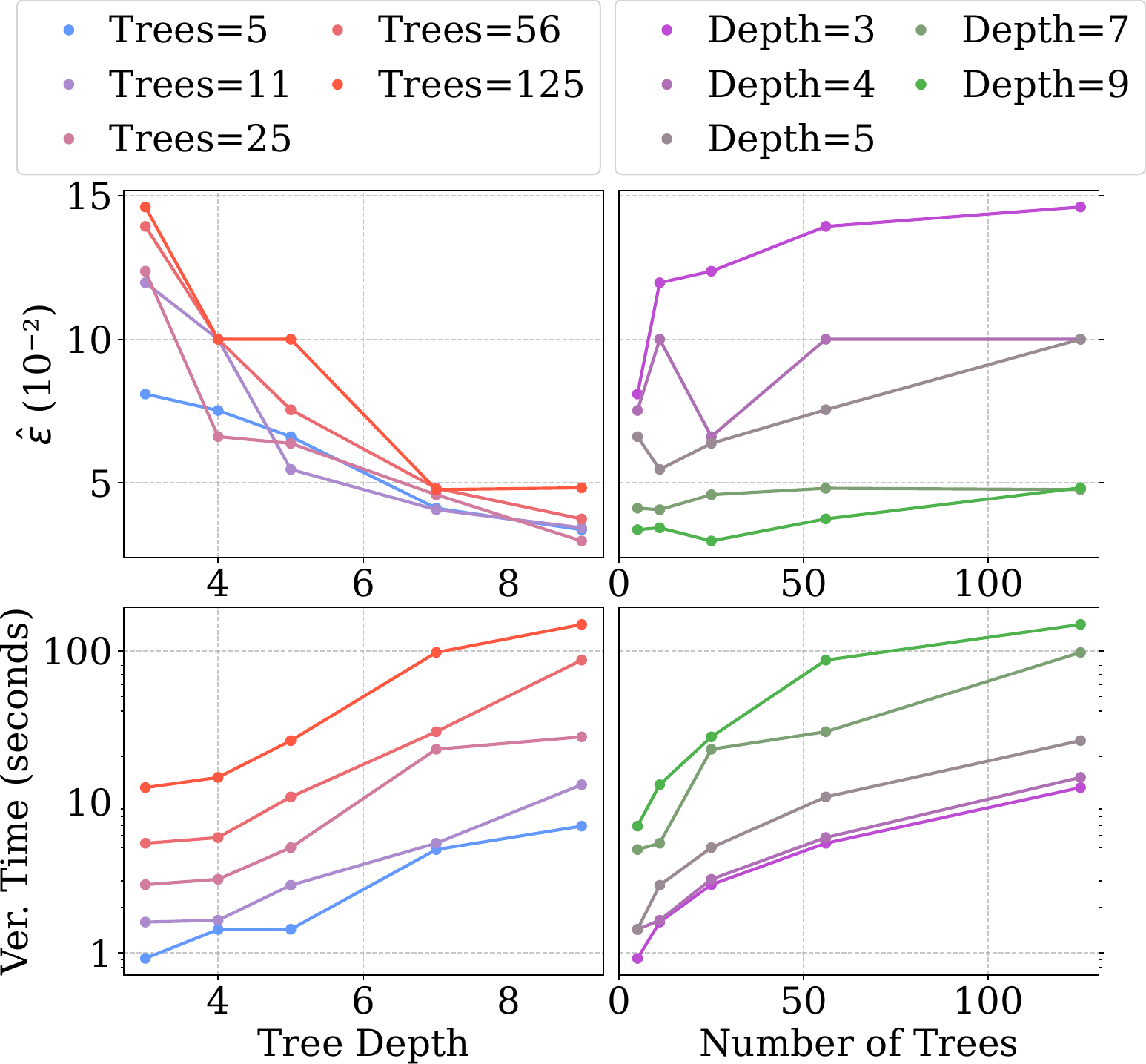}
    \subcaption[GCD Perturbation Size Search Result and Verification Time over Model Size]{on dataset GCD}
    \label{fig:size-medians/gcd}
    \end{subfigure}

    \begin{subfigure}{0.49\linewidth}
    \centering
    \includegraphics[width=\linewidth]{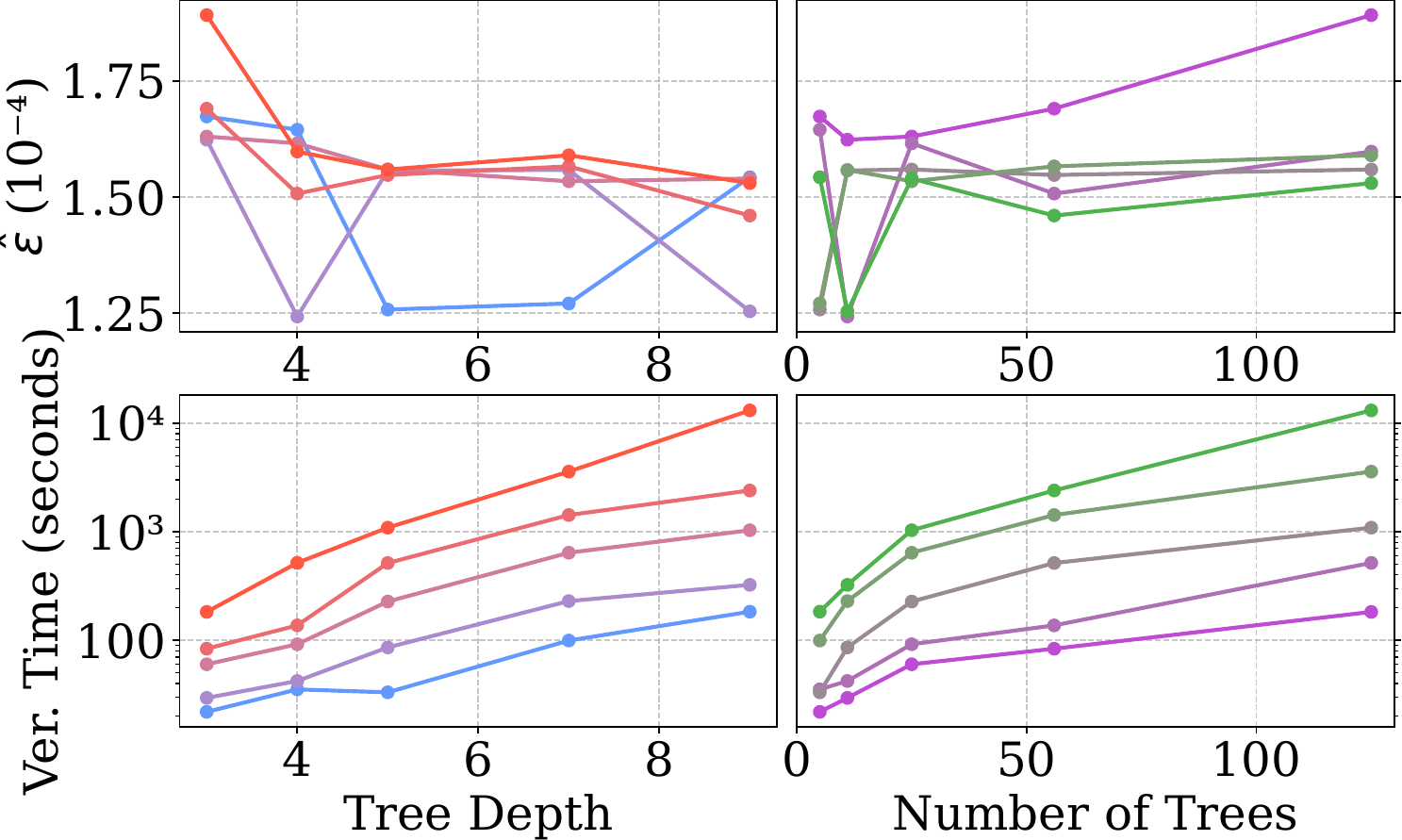}
    \subcaption[CIS Fraud Perturbation Size Search Result and Verification Time over Model Size]{on dataset IEEE CIS Fraud Prediction}
    \label{fig:size-medians/cis_fraud}
    \end{subfigure}
    \begin{subfigure}{0.49\linewidth}
    \centering
    \includegraphics[width=\linewidth]{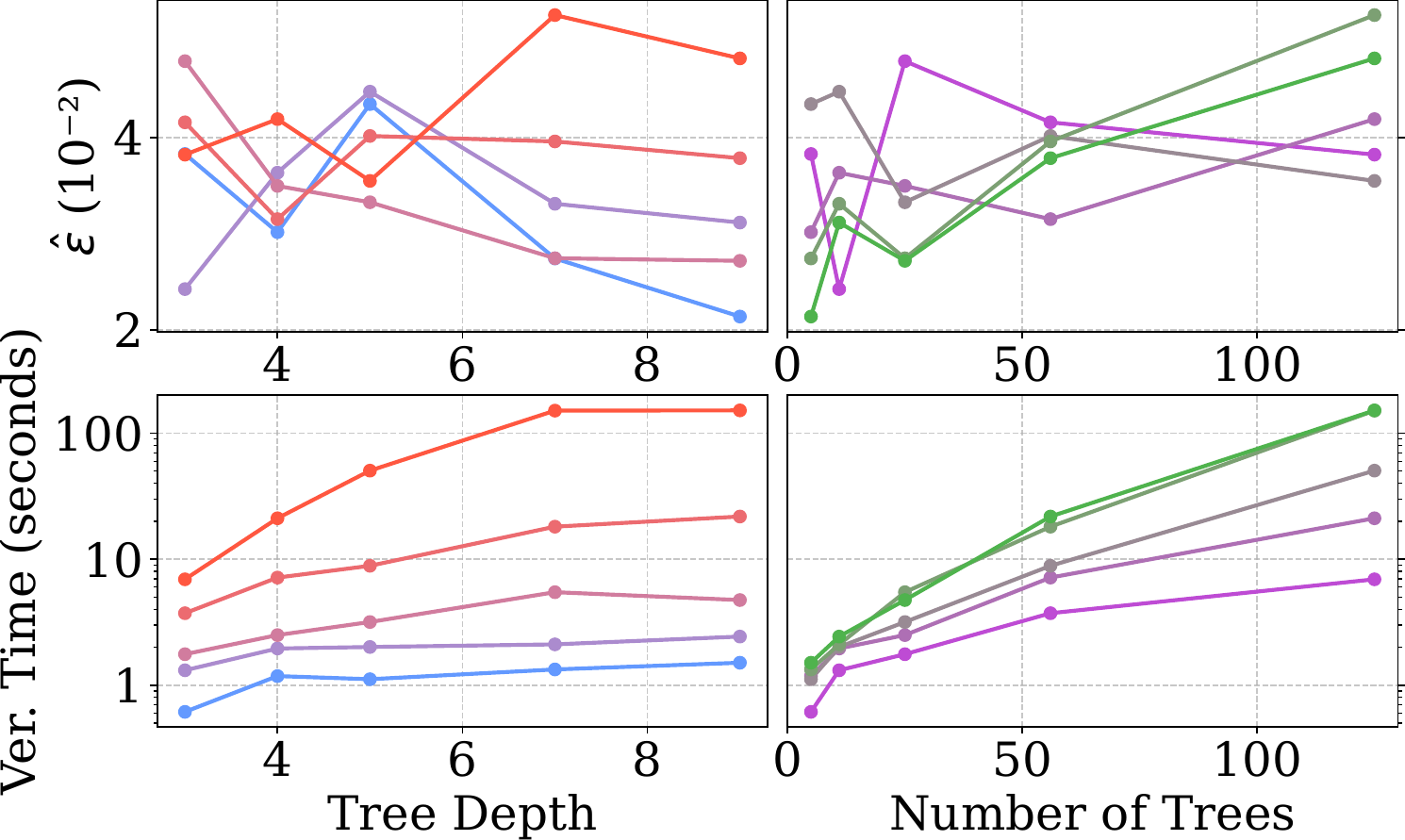}
    \subcaption[Ionosphere Perturbation Size Search Result and Verification Time over Model Size]{on dataset Ionosphere}
    \label{fig:size-medians/ionosphere}
    \end{subfigure}

    \caption{Estimated perturbation size $\hat{\varepsilon}$ and verification time spent as functions of tree depth and the number of trees; verification time is plotted on a logarithmic axis.}
    \label{fig:size-medians}
\end{figure}

\hh{modified -- check:}
We show line plots of the median values for the perturbation size determined by our algorithm and the required verification time on four of our datasets: Adult (\Cref{fig:size-medians/adult}), GCD (\Cref{fig:size-medians/gcd}), CIS Fraud (\Cref{fig:size-medians/cis_fraud}) and Ionosphere (\Cref{fig:size-medians/ionosphere}). The reported time only considers the time spent solving the MILP with SCIP, because the search algorithm introduces only minimal overhead.
\hh{briefly mention why this makes sense.}
    
As expected, we observe that verification time grows exponentially or with some large polynomial in both tree depth and the number of trees.
The estimated perturbation size seems to decrease for deeper trees on large or sufficiently complex datasets (Adult, GCD and CIS Fraud). On small or low-complexity datasets (Ionosphere, \Cref{fig:size-medians/ionosphere}), no clear relationship to tree depth is observed. 
{A possible explanation is that with such small datasets, small trees are able to learn enough and further splits are not meaningful.}
This would mean that deeper trees, trained conventionally, are more vulnerable to small perturbations.

\hh{modified:}
The influence of the number of trees on  robustness is unclear, but generally, this hyperparameter seems to have low to no effect as soon as a sufficient number of trees is trained. 
{This leans towards a different direction, which we do not explore further in the present paper, regarding the robustness of ensembles of models against their single base model and how the size of an ensemble impacts robustness.}

\hh{modified (was quite awkward and hard to understand):}
To obtain good estimates of perturbation size with minimal computational effort, meaning smaller models; it would likely be necessary to search with several configurations of the RF model. 
This could potentially result in worse scaling on easy-to-verify datasets compared to using a large models directly. 
In this context, it would also be useful to limit the maximum RF size to the smallest values that  still result in 'sufficient' performance.
From our results on hard-to-verify datasets, it seems possible to drastically lower the number of trees while tree depth would need to stay close to the final expected depth. If shallow trees have to be used, estimations for the suitable perturbation size of a deep tree would have to be obtained, likely based on sampling several smaller depth values {and extrapolating the curve to the expected depth}.
Using only a tenth of the trees in the random forests achieves 10- to over 1000-fold efficiency improvements.
These results suggest that this method is mainly applicable to hard-to-verify datasets. The area where improvement is most necessary.

To choose the value of $\hat{\varepsilon}$ for the next step of our pipeline, we optimised the hyperparameters of the random forest using a random search.
\Cref{table:epsilons} lists the perturbation sizes thus determined, which mostly align with the largest model configurations from our grid. This suggests that the best accuracy is achieved using highly complex models and that using smaller models as described above would allow for substantial efficiency gains.

\section{Efficient Training Methods}
\label{sec:exp/train}
While it has been shown that from the defensive methods, Groot is the most efficient~\cite{vos_efficient_2021}, we provide a new perspective. We do not choose the perturbation size for training to be equal to the attack size to be verified against, but optimise it 
\hh{modified -- check:}
jointly with other hyperparameters.

\subsection{Setup}
\label{sec:exp/train/setup}
The $\hat{\varepsilon}$ used in the next steps of the pipeline is chosen separately from the experiment above. As explained above, we first optimised the hyperparameters of the RFs using a random search
before using the RFs in our search procedure. This guarantees the best possible performance and thus a realistic threat scenario. On each dataset, 10 repetitions were performed; the mean values and standard deviations of the dataset-specific perturbation sizes 
are reported in \Cref{table:epsilons}.
\begin{table}[tb]
    \centering
    \begin{tabular}{r c c c l}
    \toprule
        Dataset       &  &  Mean   & & Stddev \\
    \midrule
        GCD (Statlog)     & & $0.06$ & &$0.02$  \\
        MS Malware  & &$0.007$  & &$0.005$  \\
        IEEE CIS Fraud & &$1.52\cdot 10^{-4}$  & &$0.03\cdot 10^{-4}$  \\
    \midrule
        Wine        & &$0.05$  & &$0.01$  \\
        Adult       & &$0.0033$  & &$9\cdot 10^{-4}$ \\
        Ionosphere  & & $0.04$ & &$0.01$  \\
    \bottomrule
    \end{tabular}
    \caption[Perturbation Distances for all Datasets]{Epsilon values $\hat{\varepsilon}$ for the datasets from \Cref{table:datasets} \hh{modified -- check:} 
    determined using our approach, based on a RF ensemble surrogate with optimised hyperparameters. The number of decimals is based on the standard deviation. The chosen $\hat{\varepsilon}$ is chosen as this mean.}
    \label{table:epsilons}
\end{table}

For the training experiment, we still performed 7 repetitions. 
To use the defensive methods to their best efficacy, we optimise their hyperparameters, including the perturbation size $\varepsilon$.
Among many possible objectives (accuracy, adversarial accuracy to varying attack sizes, training or inference time), we decided to maximize 
\hh{modified -- check:}
standard accuracy and adversarial accuracy for $\hat{\varepsilon}$. This way, we account for the potential accuracy gap while optimising the models to be robust.
To minimise the impact of the optimisation on total training time, we used multi-objective ParEGO~\cite{knowles_parego_2006,davins-valldaura_parego_2017}; the different objectives were balanced natively via ParEGOs one-dimensional internal representation. 
The reported training times are solely the time it took to train the optimised classifier, not the entire time required for optimisation.

The impact of adversarial verification on hyperparameter optimisation time is large, since each model trained during optimisation needs to be verified. We therefore only validated a small subsample of the data, the validation set, using the work of Kantchelian {\it et al.}~\cite{kantchelian_evasion_2016}. Preliminary results with the heuristic verification technique of Zhang {\it et al.}~\cite{zhang_efficient_2020} showed inconsistent results.

Groot RFs, Robust RFs, Treant RFs and Robust Boost inherently only allow binary targets. We use OneVsRest classifiers from sklearn for these methods on the Wine dataset, following the work of Vos and Verwer~\cite{vos_efficient_2021}. The hyperparameters 
are {trained jointly and thus} shared among the classifiers. 
For Noisy RFs, the noise used for generating the data is 
distributed uniformly at random. 


\subsection{Results}
\label{sec:exp/train/results}
\begin{figure}[tb]
\begin{subfigure}{\linewidth}
    \def\svgwidth{\linewidth}
    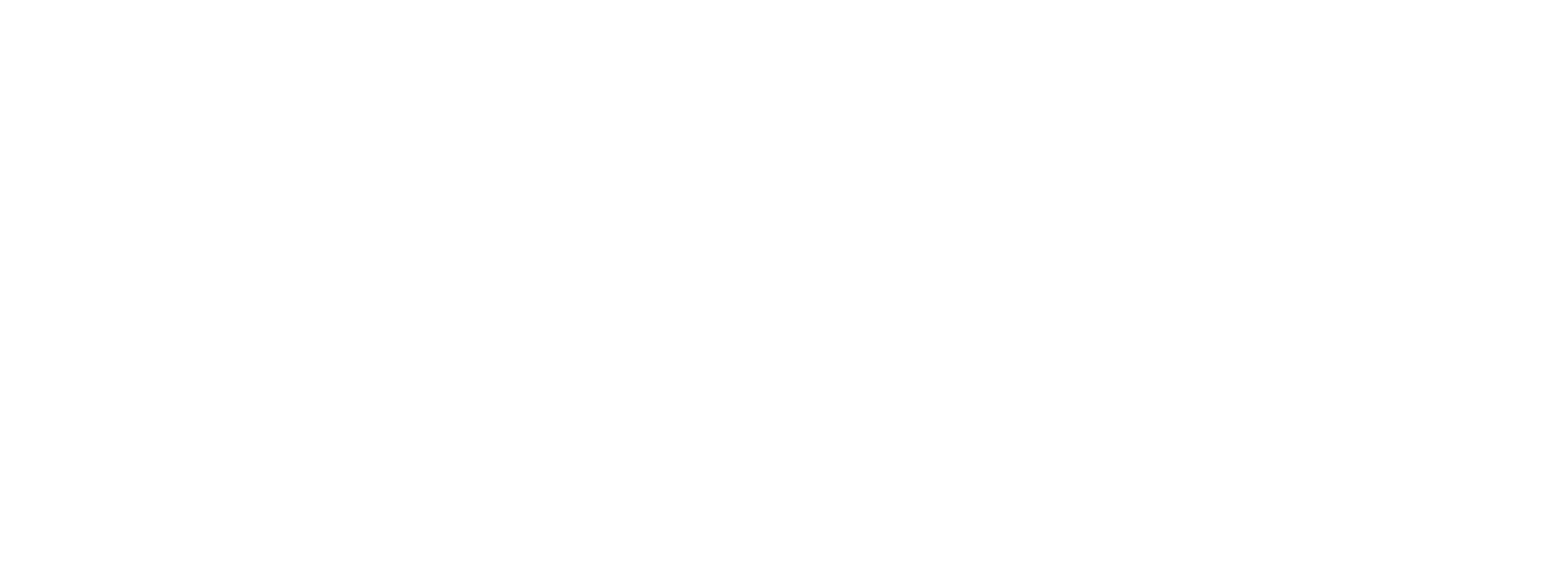
    \subcaption[CIS Fraud Training with Objective Acc+AdvAcc($\hat{\varepsilon}$)]{on IEEE CIS Fraud Detection}
    \label{fig:training/cis_fraud}
\end{subfigure}
    
\begin{subfigure}{\linewidth}
    \def\svgwidth{\linewidth}
    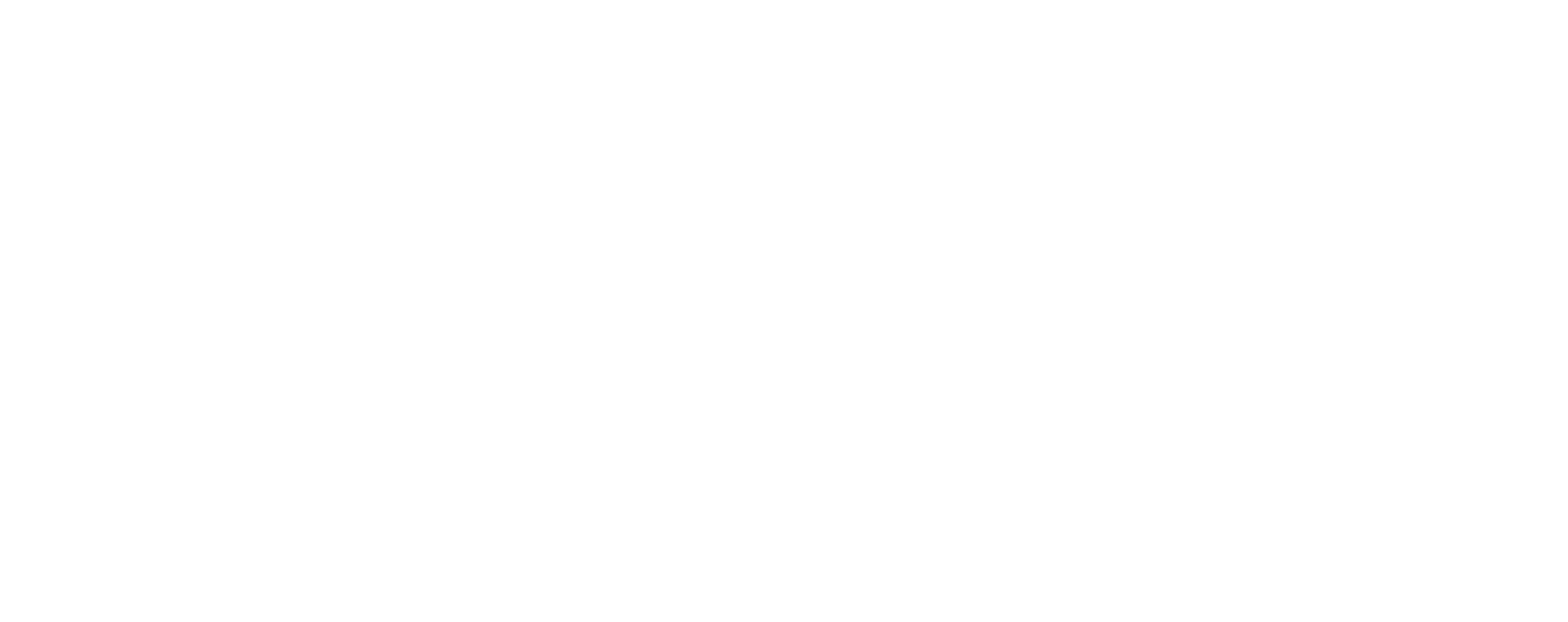
    \subcaption[Ionosphere Training with Objective Acc+AdvAcc($\hat{\varepsilon}$)]{on Ionosphere}
    \label{fig:training/ionosphere}
\end{subfigure}
    \caption{Accuracy, adversarial accuracy for $\hat{\varepsilon}$-sized perturbations and training time for all models.}
\end{figure}

For each dataset, we look at the accuracy, training time and adversarial accuracy against perturbations of size $\hat{\varepsilon}$. 
In \Cref{table:train_results}, 
we show detailed experimental results on IEEE Cis Fraud Detection (\Cref{fig:training/cis_fraud}) and Ionosphere (\Cref{fig:training/ionosphere}).
Treant RFs consistently ran out of time during training on CIS Fraud and other large datasets; for Noisy RFs, we encountered other issues.
\begin{table}[tb]
    \centering
    \label{table:train_results}
    \caption[Results of Training: Accuracy, Adv. Accuracy and Training Time]{Median accuracy, adversarial accuracy for $\hat{\varepsilon}$ and training time in seconds of all methods and datasets. RB and RT represent Robust Boost and Robust Trees, respectively.}
    \begin{tabularx}{\textwidth}{r r YYYYYYYY}
    \toprule
       Dataset  & Metric & RF & GBT & Groot& Robust& Noisy& Treant & RB & RT \\
    \midrule
        \multirow[t]{3}{*}{Wine} & Acc.       & $0.97$ & $0.93$            & $0.97$ & $0.95$            & $\textbf{0.98}$ & $0.91$ & $0.94$ & $0.95$ \\
                          & Adv.Acc.   & $0.86$            & $0.84$            & $0.86$            & $\textbf{0.88}$ & $0.84$ & $0.77$ & $0.83$ & $0.87$ \\
                          & Train Time & $\textbf{0.08}$   & $0.10$ & $0.40$            & $0.48$            & $0.09$ & $539.3$ & $11.28$ & $0.24$ \\
\addlinespace
\multirow[t]{3}{*}{MS Malware} & Acc.       & $0.58$ & $0.56$            & $0.59$           & $\textbf{0.60}$ & –      & –      & $\textbf{0.60}$ & $0.59$ \\
                          & Adv.Acc.   & $0.41$            & $0.45$            & $0.58$ & $0.57$            & –      & –      & $\textbf{0.60}$ & $0.46$ \\
                          & Train Time & $\textbf{0.62}$   & $1.86$ & $3.20$            & $4.06$            & –      & –      & $442.3$ & $14.05$ \\
\addlinespace
\multirow[t]{3}{*}{Ionosphere} & Acc.       & $0.92$ & $0.85$            & $\textbf{0.94}$ & $\textbf{0.94}$ & $0.92$ & $0.91$ & $0.92$ & $0.91$ \\
                          & Adv.Acc.   & $0.83$            & $0.80$            & $\textbf{0.92}$ & $0.88$            & $0.88$ & $0.88$ & $0.90$ & $0.80$ \\
                          & Train Time & $\textbf{0.15}$   & $0.19$ & $0.53$            & $0.28$            & $0.33$ & $2905$ & $36.4$ & $0.29$ \\
\addlinespace
\multirow[t]{3}{*}{GCD}        & Acc.       & $0.73$            & $0.69$            & $0.70$            & $0.74$ & $\textbf{0.78}$ & $0.69$ & $0.73$ & $0.73$ \\
                          & Adv.Acc.   & $0.70$            & $0.66$            & $0.69$            & $0.71$ & $0.50$ & $0.66$ & $\textbf{0.72}$ & $0.70$ \\
                          & Train Time & $\textbf{0.04}$   & $0.22$ & $0.57$            & $0.15$            & $1.71$ & $1443$ & $15.42$ & $0.37$ \\
\addlinespace
\multirow[t]{3}{*}{CIS Fraud} & Acc.       & $0.77$ & $0.76$            & $0.77$ & $0.76$            & –      & –      & $\textbf{0.78}$ & $0.74$ \\
                          & Adv.Acc.   & $0.10$            & $0.03$            & $0.76$ & $0.73$            & –      & –      & $\textbf{0.78}$ & $0.41$ \\
                          & Train Time & $\textbf{0.32}$   & $2.20$ & $2.51$            & $2.20$            & –      & –      & $2823$ & $3.14$ \\
\addlinespace
\multirow[t]{3}{*}{Adult}       & Acc.       & $\textbf{0.85}$ & $0.81$ & $\textbf{0.85}$ & $\textbf{0.85}$ & –      & –      & $\textbf{0.85}$ & $\textbf{0.85}$ \\
                          & Adv.Acc.   & $0.84$            & $0.81$ & $0.84$            & $\textbf{0.85}$ & –      & –      & $\textbf{0.85}$ & $\textbf{0.85}$ \\
                          & Train Time & $\textbf{0.27}$   & $0.35$ & $0.63$            & $0.89$            & –      & –      & $286.5$ & $0.35$ \\
    \bottomrule
    \end{tabularx}
\end{table}

Overall, GBT seems to have significantly lower accuracy as well as adversarial accuracy, 
\hh{modified -- check:}
despite requiring training time around the average over the methods we considered. 
Other methods are similar to each other regarding accuracy, none is clearly better than the others.
The training times range from 0.1 to 1000s, with fast adversarial training methods taking between 0.1 and 10s. 

Compared to a simple RFs, the adversarial defence methods Groot RF and Robust RF consistently show increased robustness. The robustness of Robust Trees and Noisy RFs appear to be dataset-dependant. They perform poorly on IEEE CIS Fraud Detection but fairly well on Ionosphere. A similar disparity in performance can also be seen on other datasets. 
Robust Boost performs particularly well, considering that it is a boosting approach like GBT, which we observed to perform worst.

Overall, Groot RFs and Robust Boost show the highest robustness, closely followed by Robust RFs and Treant RFs. However, among those well-performing methods, the training times of Robust Boost and Treant RFs are 2-3 magnitudes and 4 magnitudes larger, respectively, than those of other methods. Default RFs take the least time, whereas the other defence methods take a similar amount of time as GBTs. 
From our data, Groot RFs and Robust RFs would be the most efficient, yielding high robustness on low training time.

One notable phenomenon that {we hoped to avoid through the use of HPO, }
is that higher robustness can be achieved easily by largely predicting the majority classes. On IEEE CIS Fraud, this seems to occur for GBTs, which lose significantly in terms of accuracy in comparison to RFs, but show improved robust accuracy. This hypothesis is backed by the wide quartile range; some models trade accuracy for robustness, while others do not.


We also notice that optimising the hyperparameters of a standard RF or GBT for robustness can be effective, in that the models thus obtained can compete against robust training on some datasets, see {\it e.g.} Wine, GCD or Adult in \Cref{table:train_results}, while still requiring low training time. This, however, depends strongly on the dataset.

On some of the other datasets, 
normal DT ensembles seem to be quite robust. The $\hat{\varepsilon}$ was chosen to induce an adversarial success rate of $\eta^*$ on normally trained models; via the incorporation of robustness into the HPO objective, the hyperparameters of RFs and GBT may still be chosen such that they aid robustness.

\section{Investigating Verification Times}
\label{sec:exp/verif}
We test the impact of earlier choices on the resulting verification time. 
To the best of our knowledge, the impact of the training method on the verification time has not been investigated thoroughly in the past, despite its potentially significant impact on the efficiency of the full pipeline.
The required training time for adversarially trained RFs was improved significantly using Groot~\cite{vos_efficient_2021}, but no method has been crafted for low verification time.  

\subsection{Setup}
\label{sec:exp/verif/setup}
We verified the eight models obtained through the training experiment described in \Cref{sec:exp/train}
\hh{modified -- check:}
using SCIP and recording its running time.

\subsection{Results}
\label{sec:exp/verif/results}

\begin{figure}[t]
    
    

    \centering
    \def\svgwidth{\linewidth}
    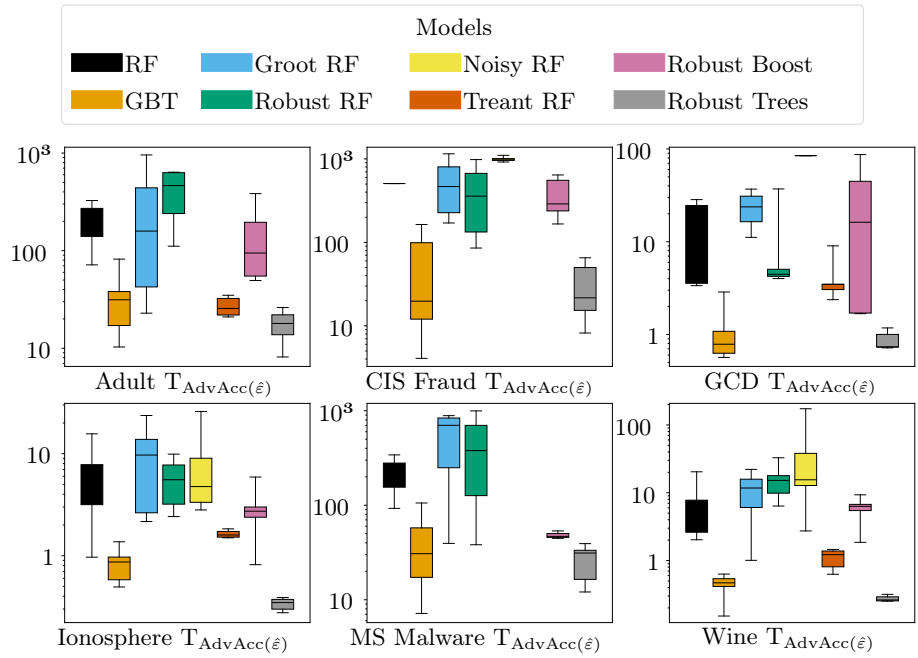
    \caption[Verification times: Boxplots for all Datasets and Methods]{Verification times in seconds on the test sets for attacks of size $\hat{\varepsilon}$ for all methods on all datasets}
    \label{fig:verif_times}
\end{figure}

For each dataset, we look at the verification time for adversarial accuracy against perturbations of size $\hat{\varepsilon}$. We show experimental results on the datasets in \Cref{fig:verif_times}.

\hh{modified:}
Verification times on our datasets lie between 0.1 and 1\,000 seconds, depending largely on the dataset. For each dataset, there is a difference of about 2 orders of magnitude between the fastest and the slowest method.

\hh{streamlined:}
Overall, the fastest models to verify are GBT and Robust Trees;
however, as shown in our earlier results, they also result in less robust models.
It seems that boosted tree ensembles are easier to verify than random forests. The Robust Boost method takes the longest within that group, but is often still faster (by up to a factor 10) than most other RF methods. 

Among the RFs, the ones constructed with Treant are generally faster to verify, taking a similar amount of time as Robust Boost. Other RF methods take roughly one order of magnitude more time than GBTs.
Verifying Noisy RFs can take a largely varying amount of time, depending on the dataset, but they are often slower to verify than other RF methods.
Default RF models take the same amount of time for verification as Groot RFs and Robust RFs (about one order of magnitude higher than GBTs). 


We note that, from the first experiment (\Cref{sec:exp/pert}), we observed that verification time grows exponentially in model size. Because we optimised the hyperparameters of the models for accuracy and adversarial accuracy in the second experiment (\Cref{sec:exp/train}), 
the hyperparameters thus determined, like tree depth and number of trees, vary between methods. It is unclear how much of the variation in verification time observed in this last set of experiments is related to the chosen tree depth and number of trees, and how much it depends on the internally constructed MILP given different splitting mechanisms. Based on preliminary investigation in the optimised hyperparameter values it seems both might be influential, but further work is needed to better understand those factors.

\section{Further Discussion}
\label{sec:exp/discussion}

The verification time does not seem to be correlated with training time, but for well-performing slow training methods ({\it e.g.}, Treant RFs and Robust Boost), we observed significantly lower verification times than for well-performing fast training methods, i.e., Groot RFs and Robust RFs.

The most efficient methods with regard to training times, Groot RFs and Robust RFs, take 10 to 100 times longer to verify than to train. On the other hand, Robust Boost is 10 times faster to verify than to train, and Treant up to 1000 times faster on Ionosphere;
however, in most cases, this still does not make up for the long training times.


Considering that, in our pipeline, during hyperparameter optimisation each model had to be both trained and verified at each 
\hh{modified -- check:}
iteration of the optimisation loop,
verification time is 
as important as training time in determining the efficiency of an adversarial training pipeline. Indeed, a shorter verification time allows many more hyperparameter values to be tested, especially for training time efficient methods.
A way to lower the cost of verification during hyperparameter optimisation could also be to use approximative verification methods, such as the one from Chen {\it et al.}~\cite{chen_robustness_2019}, or to introduce racing into the verification process, and thus only verifying on a smaller portion of validation inputs.

Our results also highlight how training methods impact verification. If this result transfers to neural network verification, which generally suffers from the practical intractability of verifying larger models, there is value in finding approaches which can train accurate and easy-to-verify networks. 

\section{Conclusion}
\label{sec:conclusion}



In this work, we have presented a preliminary empirical analysis of the computational efficiency of trustworthy DT-based machine learning models. We divided the process of training such model into three steps and investigated each stage separately, then drew high-level conclusions {regarding how these stages interact with each others.}

\paragraph{Selecting the perturbation size.} Using a simple algorithm to automatically choose the perturbation size to be used on our 6 datasets, we investigated the influence of model size on the resulting perturbation size and verification time. We found clear indications that both tree depth and the number of trees cause exponential growth in the required verification time. The robustness seems to often decrease with deeper RF models, while for the number of trees no clear relation is visible. The robustness of models on small datasets does not seem to exhibit a clear pattern either. 
\hh{slightly modified:}
For those, however, the computational effort required is relatively modest, rendering standard perturbation size estimation acceptable. For larger, complex datasets using smaller models presents a viable option, 
\hh{slightly reworded:}
potentially achieving efficiency improvements
of factors between 10 and over 1000. Future work needs to prove this strategy empirically.

\paragraph{Optimising and training robust models.} We optimised the hyperparameters of eight models to perform best with regard to accuracy and adversarial accuracy. Then, we repeated a similar analysis as found in previous work on the performance of the trained model regarding accuracy, adversarial accuracy and training time. We found the most efficient methods to be Groot RFs~\cite{vos_efficient_2021} and Robust RFs~\cite{chen_robust_2019}.
Accounting for HPO to select the training perturbation size, we were able to confirm previous results on the performance and training time efficiency of Treant RFs and Robust Boost. We are unsure why Robust Trees did not yield any robustness improvement in our experiments. 

\paragraph{Certifying robustness.} We linked the optimised training methods to the verification times of the models thus obtained. We found that gradient-boosted trees are by far easier to verify than random forests. We believe that the optimized hyperparameters alone cannot explain this result, but different solution spaces for the MILP likely also play a role.
Our results also seem to indicate that the training time and verification time of adversarial trained models are at odds with each other. Future work would be needed to confirm this, since we see no reason to believe that there is a trade-off between the two.

Overall, we have shown {that each decision taken when training trustworthy decision tree ensemble models impacts not only its performance but also its efficiency. Our findings invite further research regarding the computational cost of each step of the pipeline and how they interact, as well as the exploration of similar questions for other types of models}.

\begin{credits}
\subsubsection{\ackname}
We gratefully acknowledge
support through an Alexander-von-Humboldt Professorship
in Artificial Intelligence held by Holger Hoos. Further, the authors would like to thank Annelot Bosman for fruitful discussions and helpful feedback.

\subsubsection{\discintname}
The authors have no competing interests to declare that are
relevant to the content of this article.

\subsubsection{Statement on the Usage of Generative AI Systems.}
We used LanguageTool and Grammarly for spelling and grammar checking and, in rare cases, ChatGPT for finding improved wording.
For creating the code of our visualisations, we used Copilot with inline suggestions and Claude 3.7 for fast prototyping as well as for preliminary data analysis.

We fully reviewed all content created by such systems and take full responsibility for any content presented in this paper.

\end{credits}
%
%
%


\hh{There are still missing page numbers (cf 1. vs 2) ...}
\bibliographystyle{splncs04}
\bibliography{bib}
\end{document}